\documentclass{article}

\usepackage{microtype}
\usepackage{graphicx}
\usepackage{subfigure}
\usepackage{booktabs} 
\usepackage{amsmath}
\usepackage{amsfonts}
\usepackage{multicol, blindtext}
\usepackage{algorithm,algorithmic}
\usepackage{enumitem}
\usepackage{balance}

\usepackage{hyperref}

\usepackage[accepted]{icml2021}

\icmltitlerunning{Target Training Does Adversarial Training Without Adversarial Samples}

\begin{document}

\twocolumn[
\icmltitle{Target Training Does Adversarial Training Without Adversarial Samples}




\begin{icmlauthorlist}
\icmlauthor{Blerta Lindqvist}{to}
\end{icmlauthorlist}
\icmlaffiliation{to}{Department of Computer Science, Aalto University, Helsinki, Finland}
\icmlcorrespondingauthor{Blerta Lindqvist}{blerta.lindqvist@aalto.fi}

\icmlkeywords{Adversarial Machine Learning, ICML}

\vskip 0.3in
]



\printAffiliationsAndNotice{}  

\begin{abstract}

Neural network classifiers are vulnerable to misclassification of adversarial samples, for which the current best defense trains classifiers with adversarial samples. However, adversarial samples are not optimal for steering attack convergence, based on the minimization at the core of adversarial attacks. The minimization perturbation term can be minimized towards $0$ by replacing adversarial samples in training with duplicated original samples, labeled differently only for training. Using only original samples, Target Training eliminates the need to generate adversarial samples for training against all attacks that minimize perturbation. In low-capacity classifiers and without using adversarial samples, Target Training exceeds both default CIFAR10 accuracy ($84.3$\%) and current best defense accuracy (below $25$\%) with $84.8$\% against CW-L$_2$($\kappa=0$) attack, and $86.6$\% against DeepFool. Using adversarial samples against attacks that do not minimize perturbation, Target Training exceeds current best defense ($69.1$\%) with $76.4$\% against CW-L$_2$($\kappa=40$) in CIFAR10.



\end{abstract}

\section{Introduction}

Neural network classifiers are vulnerable to malicious adversarial samples that appear indistinguishable from original samples~\cite{szegedy2013intriguing}, for example, an adversarial attack can make a traffic stop sign appear like a speed limit sign~\cite{eykholt2018robust} to a classifier. An adversarial sample created using one classifier can also fool other classifiers~\cite{szegedy2013intriguing, biggio2013evasion}, even ones with different structure and parameters~\cite{szegedy2013intriguing, goodfellow6572explaining, papernot2016transferability, tramer2017space}. This transferability of adversarial attacks~\cite{papernot2016transferability} matters because it means that classifier access is not necessary for attacks. The increasing deployment of neural network classifiers in security and safety-critical domains such as traffic~\cite{eykholt2018robust}, autonomous driving~\cite{amodei2016aisafety}, healthcare~\cite{faust2018deep}, and malware detection~\cite{cui2018detection} makes countering adversarial attacks important.

Most current attacks, including the strongest Carlini\&Wagner attack (CW)~\cite{carlini2017towards}, are gradient-based attacks. Gradient-based attacks use the classifier gradient to generate adversarial samples from non-adversarial samples. Gradient-based attacks minimize the sum of classifier adversarial loss and perturbation~\cite{szegedy2013intriguing}, though attacks can relax the perturbation minimization to allow for bigger perturbations. The CW attack~\cite{carlini2017towards} uses the $\kappa>0$ parameter to control perturbation, while Projected Gradient Descent (PGD)~\cite{kurakin2016adversarial,madry2017towards} and FastGradientMethod (FGSM)~\cite{goodfellow6572explaining} use an $\epsilon$ parameter. Other gradient-based adversarial attacks include DeepFool~\cite{moosavi2016deepfool}, Zeroth order optimization (ZOO)~\cite{chen2017zoo}, Universal Adversarial Perturbation (UAP)~\cite{moosavi2017universal}.

Many recent proposed defenses have been broken~\cite{carlini2016defensive,carlini2017adversarial,carlini2017magnet,athalye2018obfuscated,tramer2020adaptive}. They fall largely into these categories: (1)~adversarial sample detection, (2)~gradient masking and obfuscation, (3)~ensemble, (4)~customized loss. Detection defenses~\cite{meng2017magnet,ma2018characterizing,li2018generative,hu2019new} aim to detect, correct or reject adversarial samples. Many detection defenses have been broken~\cite{carlini2017magnet,carlini2017adversarial,tramer2020adaptive}. Gradient obfuscation is aimed at preventing gradient-based attacks from access to the gradient and can be achieved by shattering gradients~\cite{guo2017countering,verma2019error,sen2020empir}, randomness~\cite{dhillon2018stochastic,li2018generative} or vanishing or exploding gradients~\cite{papernot2016distillation,song2017pixeldefend,samangouei2018defense}. Many gradient obfuscation methods have also been successfully defeated~\cite{carlini2016defensive,athalye2018obfuscated,tramer2020adaptive}. Ensemble defenses~\cite{tramer2017ensemble,verma2019error,pang2019improving,sen2020empir} have also been broken~\cite{carlini2016defensive,tramer2020adaptive}, unable to even outperform their best performing component. Customized attack losses defeat defenses~\cite{tramer2020adaptive} with customized losses~\cite{pang2019rethinking,verma2019error} but also, for example ensembles~\cite{sen2020empir}. Even though it has not been defeated, Adversarial Training~\cite{szegedy2013intriguing,kurakin2016adversarial,madry2017towards} assumes that the attack is known in advance and generates adversarial samples at every training iteration. The inability of recent defenses to counter adversarial attacks calls for new kinds of defensive approaches.

In this paper, we make the following major contributions:
\setlist{nolistsep}
\begin{itemize}[noitemsep]

\item We develop Target Training - a novel, white-box defense that eliminates the need to know the attack or to generate adversarial samples in training for attacks that minimize perturbation.

\item Target Training accuracy in CIFAR10 is $84.8$\% against CW-L$_2$($\kappa=0$) and $86.6$\% against DeepFool. This even exceeds CIFAR10 default accuracy ($84.3$\%).

\item Using adversarial samples against attacks that do not minimize perturbation, Target Training accuracy against CW-L$_2$($\kappa=40$) in CIFAR10 is $76.4$\%, which exceeds Adversarial Training accuracy $69.1$\%.

\item Contrary to prior work~\cite{madry2017towards}, we find that low-capacity classifiers can counter non-$L_\infty$ attacks successfully.

\item Target Training even improves default classifier accuracy ($84.3$\%) on non-adversarial samples in CIFAR10 with $86.7$\%.

\item Our work questions whether Adversarial Training defense works by populating sparse areas, since Target Training is a form of Adversarial Training that successfully uses only original samples against attacks that minimize perturbation.

\item Contrary to~\citet{carlini2017towards}, but in support of~\citet{kurakin2018adversarial}, our experiments show that targeted attacks are much weaker than untargeted attacks.

\end{itemize}

\section{Background And Related Work}
\label{rel_work}
Here, we present the state-of-the-art in adversarial attacks and defenses, as well as a summary.

\textbf{Notation} A $k$-class neural network classifier that has $\theta$ parameters is denoted by a function $f(x)$ that takes input $x\in\mathbb{R}^{d}$ and outputs $y\in\mathbb{R}^{k}$ , where $d$ is the dimensionality and $k$ is the number of classes. An adversarial sample is denoted by $x_{adv}$. Classifier output is $y$, where $y_i$ is the probability that the input belongs to class~$i$. Norms are denoted as $L_0$, $L_2$ and $L_{\infty}$.

\subsection{Adversarial Attacks}
\label{sec-adv-attacks}

\textbf{Targeted Attacks}~\citet{szegedy2013intriguing} were the first to formulate the generation of adversarial samples as a constrained minimization of the perturbation under an $L_p$ norm. Because this formulation can be hard to solve, \citet{szegedy2013intriguing} reformulated the problem as a gradient-based, two-term minimization of the weighted sum of perturbation and classifier loss. For targeted attacks, this minimization is:

\begin{align} \label{eq_targeted_attacks}
& {\text{minimize}}
& &  c \cdot \| \delta \| + loss_{f}(x+\delta, l)  \\
& \text{such that} \nonumber
& &  x+\delta \in [0,1]^d,
\end{align}

where $c$ is a constant, $\delta$ is perturbation, $f$ is the classifier, $loss_{f}$ is classifier loss, $l$ is an adversarial label. $\| \delta \|$ in term (1) of Minimization~\ref{eq_targeted_attacks} is a norm of the adversarial perturbation, while term (2) is there to utilize the classifier gradient to find adversarial samples that minimize classifier adversarial loss. By formulating the problem of finding adversarial samples this way, \citet{szegedy2013intriguing} paved the way for adversarial attacks to utilize classifier gradients in adversarial attacks.

~\ref{eq_targeted_attacks} is the foundation for many gradient-based attacks, though many tweaks can and have been applied. Some attacks follow~\ref{eq_targeted_attacks} implicitly~\cite{moosavi2016deepfool}, and others explicitly~\cite{carlini2017towards}. The type of $L_p$ norm in term (1) of the minimization also varies. For example the CW attack~\cite{carlini2017towards} uses $L_0$, $L_2$ and $L_\infty$, whereas DeepFool~\cite{moosavi2016deepfool} uses the $L_2$ norm. A special perturbation case is the Pixel attack by~\citet{su2019one} which changes exactly one pixel. Some attacks even exclude term (1) from the~\ref{eq_targeted_attacks} and introduce an external parameter to control perturbation. The FGSM attack by~\citet{goodfellow6572explaining}, for example, uses an $\epsilon$ parameter, while the CW attack~\cite{carlini2017towards} uses a $\kappa$ confidence parameter.

There are three ways~\citep{carlini2017towards,kurakin2018adversarial} to choose what the target adversarial label is: (1)~\emph{Best case} - try the attack with all adversarial labels and choose the label that was the easiest to attack; (2)~\emph{Worst case} - try the attack with all adversarial labels and choose the label that was the toughest to attack; (3)~\emph{Average case} - choose a target label uniformly at random from the adversarial labels.

\textbf{Untargeted Attacks} Untargeted attacks aim to find a nearby sample that misclassifies, without aiming for a specific adversarial label. Some untargeted attacks, such as DeepFool and UAP, have no targeted equivalent.

\textbf{Stronger Attacks} There are conflicting accounts of which attacks are stronger, targeted attacks or untargeted attacks. \citet{carlini2017towards} claim that targeted attacks are stronger. However, \citet{kurakin2018adversarial} find targeted attacks, including worst-case targeted attacks, to be much weaker than untargeted attacks. 

\textbf{Fast Gradient Sign Method} The Fast Gradient Sign Method by~\citet{goodfellow6572explaining} is a simple, $L_{\infty}$-bounded attack that constructs adversarial samples by perturbing each input dimension in the direction of the gradient by a magnitude of $\epsilon$: $x_{adv}=x+\epsilon \cdot sign(\nabla_x loss(\theta,x,y))$.


\textbf{CarliniWagner} The current strongest attack is CW~\cite{carlini2017towards}. CW customizes~\ref{eq_targeted_attacks} by passing $c$ to the second term, and using it to tune the relative importance of the terms. With a further change of variable, CW obtains an unconstrained minimization problem that allows it to optimize directly through back-propagation. In addition, CW has a $\kappa$ parameter for controlling the confidence of the adversarial samples. For $\kappa>0$ and up to $100$, the CW attack allows for more perturbation in the adversarial samples it generates.


\textbf{DeepFool} The DeepFool attack by~\citet{moosavi2016deepfool} is an untargeted attack that follows the~\ref{eq_targeted_attacks} implicitly, finding the closest, untargeted adversarial sample. DeepFool~\cite{moosavi2016deepfool} looks at the smallest distance of a point from the classifier decision boundary as the minimum amount of perturbation needed to change its classification. Then DeepFool approximates the classifier with a linear one, estimates the distance from the linear boundary, and then takes steps in the direction of the closest boundary until an adversarial sample is found.

\textbf{Black-box attacks} Black-box attacks assume no access to classifier gradients. Such attacks with access to output class probabilities are called score-based attacks, for example the ZOO attack~\cite{chen2017zoo}, a black-box variant of the CW attack~\cite{carlini2017towards}.
Attacks with access to only the final class label are decision-based attacks, for example the Boundary~\cite{brendel2017decision} and the HopSkipJumpAttack~\cite{chen2019hopskipjumpattack} attacks.

\textbf{Multi-step attacks} The PGD attack~\cite{kurakin2016adversarial} is an iterative method with an $\alpha$ parameter that determines a step-size perturbation magnitude. PGD starts at a random point $x_0$ and then projects the perturbation on an $L_p$-ball $B$ at each iteration: ${x(j+1)=Proj_B(x(j)+\alpha \cdot sign(\nabla_x loss(\theta,x(j),y))}$. The BIM attack~\cite{kurakin2016adversarial} applies FGSM~\cite{goodfellow6572explaining} iteratively with an $\alpha$ step. To find a universal perturbation, UAP~\cite{moosavi2017universal} iterates over the images and aggregates perturbations calculated as in DeepFool.




\subsection{Adversarial Defenses}

\textbf{Adversarial Training.} Adversarial Training~\cite{szegedy2013intriguing,kurakin2016adversarial,madry2017towards} is one of the first and few, undefeated defenses. It defends by populating low probability, so-called blind spots~\cite{szegedy2013intriguing,goodfellow6572explaining} with adversarial samples labelled correctly, redrawing boundaries. The drawback of Adversarial Training is that it needs to know the attack in advance, and it needs to generate adversarial samples during training. The Adversarial Training Algorithm~2 in the Appendix is based on~\citet{kurakin2016adversarial}. \citet{madry2017towards} formulate their defense as a robust optimization problem, and use adversarial samples to augment the training. Their solution however necessitates high-capacity classifiers - bigger models with more parameters.


\textbf{Detection defenses} Such defenses detect adversarial samples implicitly or explicitly, then correct or reject them. So far, many detection defenses have been defeated. For example, ten diverse detection methods (other network, PCA, statistical properties) were defeated with attack loss customization by~\citet{carlini2017adversarial};~\citet{tramer2020adaptive} used attack customization against~\citet{hu2019new}; attack transferability~\cite{carlini2017magnet} was used against MagNet by~\citet{meng2017magnet}; deep feature adversaries~\cite{sabour2015adversarial} against~\citet{roth2019odds}.

\textbf{Gradient masking and obfuscation} Many defenses that mask or obfuscate the classifier gradient have been defeated~\cite{carlini2016defensive,athalye2018obfuscated}. \citet{athalye2018obfuscated} identify three types of gradient obfuscation: 
(1) Shattered gradients - incorrect gradients caused by non-differentiable components or numerical instability, for example with multiple input transformations by~\citet{guo2017countering}. \citet{athalye2018obfuscated} counter such defenses with Backward Pass Differentiable Approximation.
(2) Stochastic gradients in randomized defenses are overcome with Expectation Over Transformation by~\citet{athalye2017synthesizing}. Examples are Stochastic Activation Pruning~\cite{dhillon2018stochastic}, which drops layer neurons based on a weighted distribution, and~\cite{xie2017mitigating} which adds a randomized layer to the classifier input.
(3) Vanishing or exploding gradients are used, for example, in Defensive Distillation (DD)~\cite{papernot2016distillation} which reduces the amplitude of gradients of the loss function. Other examples are PixelDefend~\cite{song2017pixeldefend} and Defense-GAN~\cite{samangouei2018defense}. Vanishing or exploding gradients are broken with parameters that avoid vanishing or exploding gradients~\cite{carlini2016defensive}.

\textbf{Complex defenses} Defenses combining several approaches, for example~\cite{li2018generative} which uses detection, randomization, multiple models and losses, can be defeated by focusing on the main defense components~\cite{tramer2020adaptive}. In particular, ensemble defenses do not perform better than their best components. \citet{verma2019error,pang2019improving,sen2020empir} are defeated ensemble defenses combined with numerical instability~\cite{verma2019error}, regularization~\cite{pang2019improving}, or mixed precision on weights and activations~\cite{sen2020empir}.


\subsection{Summary} Many defenses have been broken. They focus on changing the classifier. Instead, our Target Training defense changes the classifier minimally, but focuses on steering attack convergence. Target Training is the first defense based on the minimization term of the~\ref{eq_targeted_attacks} at the core of untargeted gradient-based adversarial attacks.

\section{Target Training}

Just as adversarial attacks have used the gradient term of \ref{eq_targeted_attacks} against defenses, the perturbation term in the same minimization can be used to steer attack convergence. By training the classifier with duplicated original samples labeled differently only in training with target labels (hence Target Training), attacks that minimize perturbation are forced to converge to benign samples because the duplicated samples minimize the perturbation towards $0$.


Target Training is a form of Adversarial Training that replaces adversarial samples with orginal samples, leading attacks to converge to non-adversarial samples as they do in Adversarial Training defense. In Target Training, the final no-weight layer used in inference and testing essentially relabels the target labels to the original labels, which is the equivalent of labeling adversarial samples correctly in Adversarial Training. However, the fact that Target Training is a form of Adversarial Training that uses no adversarial samples against attacks that minimize perturbation presents us with a question: Might it be that Adversarial Training works not because it populates the distribution blind spots with adversarial samples, but because these adversarial samples steer attack convergence?

Target Training could also be extended to defend against more than one attack at the same time. For example, to defend simultaneously against two types of attacks that do not minimize perturbation, the batch size would be tripled and the batch would be populated with adversarial samples from both attacks. In addition, there would be two sets of target labels, one set for each attack.

\subsection{Classifier Architecture}
\label{sec-classif-arch}

We choose low-capacity classifiers in order to investigate whether Target Training can defend such classifiers. The MNIST classifier has two convolutional layers with 32 and 64 filters respectively, followed each by batch normalization, then a 2 × 2 max-pooling layer, a drop-out layer, a fully connected layer with 128 units, then another drop-out layer, then a softmax layer with 20 outputs, then a summation layer without weights that adds up the softmax outputs two-by-two and has 10 outputs. The CIFAR10 classifier has 3 groups of layers, each of which has: two convolutional layers with increasing number of filters and with elu activation followed by batch normalization, then a 2 × 2 max-pooling layer, then a drop-out layer. Then a softmax layer with 20 outputs, and finally a no-weight summation layer that takes softmax layer outputs as inputs, sums them two-by-two, and has 10 outputs. Table~7 in the Appendix shows classifier architecture for CIFAR10 and MNIST in detail. Training uses all layers up to the softmax layer, but not the final layer. Inference and testing uses all classifier layers.

\subsection{Training}

\begin{figure}[tb]
\vskip 0.2in
\begin{center}
\centerline{\includegraphics[width=0.8\columnwidth]{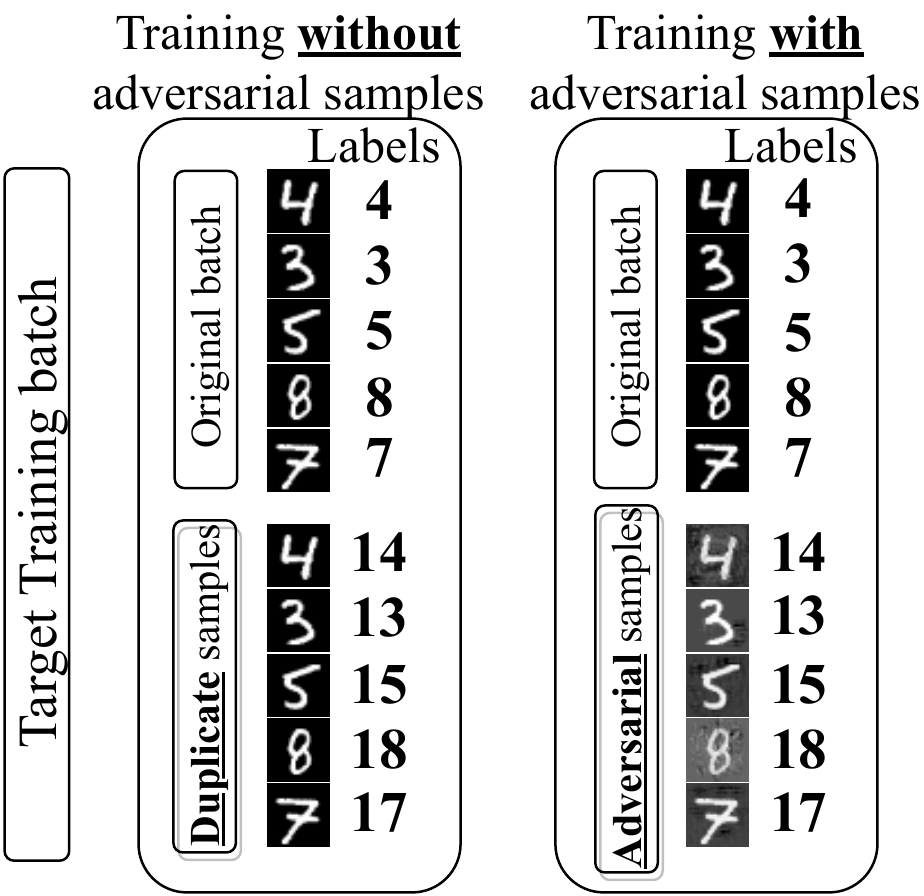}}
\caption{Outline of Target Training training without adversarial samples against attacks that minimize perturbation, and with adversarial samples against attacks that do not minimize perturbation.}
\label{tt-training}
\end{center}
\vskip -0.2in
\end{figure}

\begin{figure}[tb]
\vskip 0.2in
\begin{center}
\centerline{\includegraphics[width=0.8\columnwidth]{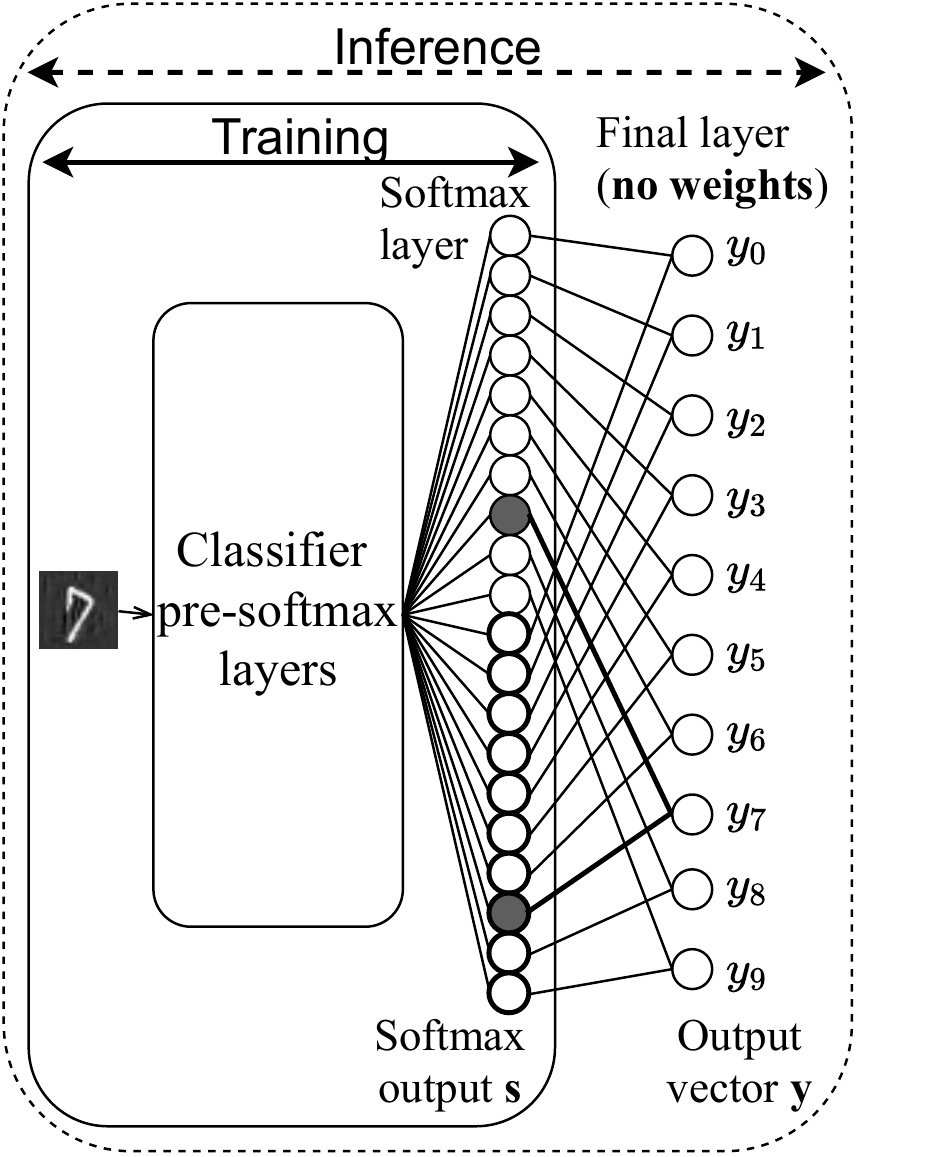}}
\caption{Outline of difference between training and inference in Target Training. All the classifier layers up to and including the softmax layer with $2k$ classes, are included in training. The final, no-weight, summation layer is not included in training, but is used in inference and testing.}
\label{tt-all}
\end{center}
\vskip -0.2in
\end{figure}

Against attacks that \emph{minimize} perturbation, such as CW(${\kappa=0}$) and DeepFool, Target Training uses duplicates of original samples in each batch instead of adversarial samples because these samples minimize the perturbation to $0$ - no other points can have smaller distance from original samples. This eliminates the overhead of calculating adversarial samples against all attacks that minimize perturbation. Figure~\ref{tt-training} shows how Target Training trains without adversarial samples to counter attacks that minimize perturbation, and with adversarial samples to counter attacks that do not minimize perturbation. Training in Target Training is also illustrated in Figure~\ref{tt-all} to show that all the layers up to the softmax layer (with $2k$ class outputs) take part in training, but not the last layer. The duplicated samples are labeled as $i+k$, where $i$ is the original label and $k$ is the number of classes.

Algorithm~\ref{alg-TT-no-perturb} shows Target Training algorithm against all attacks that minimize perturbations. Against attacks that \emph{do not minimize} perturbation, such as CW($\kappa>0$), PGD and FGSM, Target Training uses adversarial samples in training, shown in Algorithm~3 in the Appendix. Both Target Training algorithms are based on the Adversarial Training~\cite{kurakin2016adversarial} Algorithm~2 in the Appendix.

\vskip 0.2in
\begin{algorithm}[tb]
\caption{Target Training of classifier $N$ against attacks that minimize perturbation based on Adversarial Training Algorithm~2 in the Appendix.}
\label{alg-TT-no-perturb}
\begin{algorithmic}
\REQUIRE $m$ batch size, $k$ classes, classifier $N$ with all layers up to softmax layer with $2k$ output classes, TRAIN	trains a classifier on a batch and labels
\ENSURE Classifier $N$ is Target-Trained against all attacks that minimize perturbation
\WHILE{training not converged}
   \STATE $B = \{x^1,..., x^m\}$\COMMENT{Get random batch}
   \STATE $G = \{y^1,..., y^m\}$\COMMENT{Get batch ground truth}
   \STATE $B' = \{x^1,..., x^m,x^1,..., x^m\}$\COMMENT{Duplicate batch}
   \STATE $G'=\{y^1,..., y^m,y^1+k,..., y^m+k\}$\COMMENT{Duplicate ground truth and increase duplicates by $k$}
   \STATE TRAIN($N$, $B'$, $G'$)\COMMENT{Train classifier on duplicated batch and new ground truth}
 \ENDWHILE
\end{algorithmic}
\end{algorithm}
\vskip -0.2in

\subsection{Inference}

Figure~\ref{tt-all} shows that inference in Target Training differs from training by using a no-weight, final layer. The final layer derives the $y_i$ output probability of the final layer, as the sum of probabilities $s_i$ and $s_{i+k}$ in the softmax layer output: ${y_i = s_i + s_{i+k}}$, where $k$ is the number of classes, ${i \in [0 \ldots (k-1)]}$, $s$ is softmax layer output, $y$ is final layer output.

\section{Experiments}

\textbf{Threat model} We assume that the \emph{adversary goal} is to generate adversarial samples that cause misclassification. We perform white-box evaluations, assuming the adversary has complete \emph{knowledge} of the classifier and how the defense works. In terms of \emph{capabilities}, we assume that the adversary is gradient-based, has access to the CIFAR10 and MNIST image domains and is able to manipulate pixels. Perturbations are assumed to be $L_p$-constrained. For attacks that do not minimize perturbations, we assume that the attack is of the same kind as the attack used to generate the adversarial samples used during training. We assume that the adversary can generate both targeted and untargeted attacks.

\textbf{Targeted and untargeted attacks} There are conflicting views~\cite{carlini2017towards,kurakin2018adversarial} whether targeted or untargeted attacks are stronger. To determine which are stronger, we conduct experiments with both untargeted, and average-case targeted attacks where the target label is chosen uniformly at random from adversarial labels. Targeted attacks are not applicable for DeepFool.

\textbf{Attack parameters} For \emph{CW}: 9 steps, $0-40$ confidence values, default $1K$ iterations, but also experiments with up to $10K$ iterations in adaptive attacks. For \emph{PGD}, parameters based on PGD paper~\cite{madry2017towards}: for CIFAR10, $7$ steps of size $2$ with a total $\epsilon=8$; for MNIST, $40$ steps of size $0.01$ with a total $\epsilon=0.3$. For all PGD attacks, we use $0$ random initialisations within the $\epsilon$ ball, effectively starting PGD attacks from the original images. For \emph{FGSM}: $\epsilon=0.3$, as in~\cite{madry2017towards}.

\textbf{Classifier models} We purposefully do not use high-capacity models, such as ResNet~\cite{he2016deep}, used for example by~\citet{madry2017towards}, to show that Target Training does not necessitate high model capacity to defend against adversarial attacks. The architectures of MNIST and CIFAR datasets are described in Subsection~\ref{sec-classif-arch} and shown in Table~7 in the Appendix. No data augmentation was used. Default accuracies without attack are $84.3$\% for CIFAR10 and $99.1$\% for MNIST. Adversarial Training and default classifiers have same architecture, except the softmax layer is the last layer and has 10 outputs.

\textbf{Datasets} The MNIST~\cite{lecun1998mnist} and the CIFAR10~\cite{krizhevsky2009cifar} datasets are $10$-class datasets that have been used throughout previous work. The MNIST~\cite{lecun1998mnist} dataset has $60K$, $28 \times 28 \times 1$ hand-written, digit images. The CIFAR10~\cite{krizhevsky2009cifar} dataset has $70K$, $32 \times 32 \times 3$ images. Each dataset has $10K$ testing samples and all experimental evaluations are done with testing samples.

\textbf{Tools} Adversarial samples generated with CleverHans~3.0.1~\cite{papernot2018cleverhans} for CW-$L_2$~\cite{carlini2017towards}, DeepFool~\cite{moosavi2016deepfool} attacks and IBM Adversarial Robustness 360 Toolbox (ART) toolbox 1.2~\cite{art2018} for CW-$L_{\infty}$~\cite{carlini2017towards}, FGSM~\cite{goodfellow6572explaining} and PGD~\cite{kurakin2016adversarial} attacks. Target Training has been written in Python~3.7.3, using Keras~2.2.4~\cite{chollet2015keras}.

\textbf{Baselines} We choose Adversarial Training as a baseline because it is the current best defense since other defenses have been defeated successfully~\cite{carlini2016defensive,carlini2017magnet,carlini2017adversarial,athalye2018obfuscated,tramer2020adaptive}, more details in Section~\ref{rel_work}. Our Adversarial Training implementation is based on~\cite{kurakin2016adversarial}, shown in Algorithm~2 in the Appendix. We choose the ~\citet{kurakin2016adversarial} implementation and not the robust optimization of \citet{madry2017towards}, because the Adversarial Training solution by \citet{madry2017towards} necessitates high-capacity classifiers. However, we do not use high-capacity classifiers in order to show that Target Training can defend low-capacity classifiers.

\subsection{Targeted Attacks Are Weaker Than Untargeted Attacks}

There are conflicting views~\cite{carlini2017towards,kurakin2018adversarial} whether targeted or untargeted attacks are stronger. ~\citet{carlini2017towards} claim that targeted attacks are stronger, whereas ~\citet{kurakin2018adversarial} claim that targeted attacks, even worst-case ones, are much weaker that untargeted attacks. Here, we aim to find out which type of attack is stronger, to use for the rest of the experiments. We use average-case targeted attacks, expained in Section~\ref{sec-adv-attacks}.

Accuracy values of default classifiers in Table~\ref{tbl-partial-targ-untarg} show targeted attacks to be not strong. For example, targeted CW-$L_2$($\kappa=0$) in CIFAR10 decreases default classifier accuracy by less than $1$\% point, whereas targeted CW-$L_2$($\kappa=40$) attack even increases default classifier accuracy. Similarly, all targeted attacks against MNIST default classifier reduce accuracy by less than $3$\%.

By comparison, each untargeted attack is much stronger than its targeted equivalent, supporting ~\citet{kurakin2018adversarial}. Untargeted CW-$L_2$($\kappa=0$) and untargeted CW-$L_2$($\kappa=40$) in CIFAR10 reduce default classifier accuracy to below $9$\%, and in MNIST to below $1$\%.

\textbf{Based on Table~\ref{tbl-partial-targ-untarg} results, we use untargeted attacks for evaluating the performance of Target Training in the following experiments.} Additional experiments with targeted attacks using Target Training and Adversarial Training are shown in Table~8 in the Appendix.

\begin{table}[ht]
\caption{ Here, we show that each untargeted attack diminishes the accuracy of default classifiers much more than its equivalent targeted attack. This indicates that untargeted attacks are stronger. DeepFool attack has no targeted equivalent.}
\label{tbl-partial-targ-untarg}
\vskip 0.15in
\begin{center}
\begin{small}
\begin{sc}
  \begin{tabular}{lrr}
    \toprule
                                & CIFAR10    & MNIST      \\
     Attack                     & Default    & Default    \\
                                & Classifier & Classifier \\
    \midrule
    No Attack                         & 84.3\%     & 99.1\% \\
    \midrule
    Targeted Attacks                  & & \\
    \midrule
    CW-$L_2$($\kappa=0$)              & 84.0\%     & 98.3\% \\ 
    CW-$L_{\infty}$($\kappa=0$)       & 72.7\%     & 98.6\% \\ 
    DeepFool                          & NA         & NA     \\ 
    CW-$L_2$($\kappa=40$)             & 85.7\%     & 99.0\% \\ 
    PGD($\epsilon=8$, $\epsilon=0.3$) & 44.9\%     & 96.4\% \\ 
    FGSM($\epsilon=0.3$)              & 46.4\%     & 96.4\% \\ 
    \midrule
    Untargeted Attacks                & & \\
    \midrule
    CW-$L_2$($\kappa=0$)              &  8.5\%    &  0.8\% \\ 
    CW-$L_{\infty}$($\kappa=0$)       & 23.6\%    & 94.2\% \\ 
    DeepFool                          &  8.6\%    &  2.8\% \\ 
    CW-$L_2$($\kappa=40$)             &  7.9\%    &  0.8\% \\ 
    PGD($\epsilon=8$, $\epsilon=0.3$) & 10.9\%    & 90.7\% \\ 
    FGSM($\epsilon=0.3$)              & 17.6\%    & 90.7\% \\ 
    \bottomrule
  \end{tabular}
\end{sc}
\end{small}
\end{center}
\vskip -0.1in
\end{table}

\subsection{Target Training Against Attacks That Minimize Perturbation}

Table~\ref{tbl-no-perturb} shows that Target Training exceeds by far accuracies by Adversarial Training and default classifier against attacks that minimize perturbation. Without using adversarial samples in training, Target Training exceeds even default accuracy on non-adversarial samples against CW-$L_2$($\kappa=0$) and DeepFool in CIFAR10. The only case where performances are roughly equal is against CW-$L_{\infty}$($\kappa=0$) in CIFAR10. 

\begin{table*}[ht]
\caption{Here, we show Target Training performance against attacks that minimize perturbation, for which Target Training does not use adversarial samples. Target Training even exceeds performance of default classifier against CW-$L_2$($\kappa=0$) and DeepFool in CIFAR10. Target Training also exceeds the performance of Adversarial Training classifier that uses adversarial samples, except for CW-$L_{\infty}$($\kappa=0$) in CIFAR10 where accuracies are roughly equal.}
\label{tbl-no-perturb}
\vskip 0.15in
\begin{center}
\begin{small}
\begin{sc}
\begin{tabular}{lrrrrrr}
    \toprule
                        & \multicolumn{3}{c}{CIFAR10 (84.3\%)} & \multicolumn{3}{c}{MNIST (99.1\%)} \\
                                                 \cmidrule(r){2-4}   \cmidrule(r){5-7}
    Untargeted        & Target       & Adversarial    & Default      & Target       & Adversarial   & Default   \\                                                 
    Attack            & Training     & Training       & Classifier   & Training     & Training      & Classifier   \\
    \midrule
    CW-$L_2$($\kappa=0$)        &  84.8\%  &  22.8\%  &  8.5\%     & 96.9\%   &  5.0\%  &  0.8\%    \\
    CW-$L_{\infty}$($\kappa=0$) &  21.3\%  &  21.4\%  & 23.6\%     & 96.1\%   & 75.8\%  & 94.2\%    \\
    DeepFool                    &  86.6\%  &  24.0\%  &  8.6\%     & 94.9\%   &  5.2\%  &  2.8\%    \\
    \bottomrule
  \end{tabular}
\end{sc}
\end{small}
\end{center}
\vskip -0.1in
\end{table*}

\subsection{Target Training Against Attacks That Do Not Minimize Perturbation} 

Table~\ref{tbl-with-perturb} shows that Target Training can even improve accuracy compared to Adversarial Training against attacks that do not minimize perturbation, for attacks CW-$L_2$($\kappa=40$) and FGSM($\epsilon=0.3$) in CIFAR10. Against such attacks, Target Training uses adversarial samples in training as Adversarial Training does. Against PGD attack, Target Training performs worse then Adversarial Training. We attribute the Target Training performance against PGD to the low capacity of the classifiers we use. Such effect of classifier capacity on performance has been previously observed by~\citet{madry2017towards}. We anticipate Target Training performance to improve for higher-capacity classifiers.

\begin{table*}[ht]
\caption{Using adversarial samples in training, Target Training performs better than Adversarial Training against non-$L_\infty$ attacks that do not minimize perturbation in CIFAR10. Against $L_\infty$ attacks, Target Training performs worse than Adversarial Training.}
\label{tbl-with-perturb}
\vskip 0.15in
\begin{center}
\begin{small}
\begin{sc}
  \begin{tabular}{lrrrrrr}
    \toprule
                            &  \multicolumn{3}{c}{CIFAR10 (84.3\%)}  & \multicolumn{3}{c}{MNIST (99.1\%)} \\
                                                                 \cmidrule(r){2-4}   \cmidrule(r){5-7}
     Untargeted        & Target       & Adversarial    & Default      & Target       & Adversarial   & Default   \\                                                 
     Attack            & Training     & Training       & Classifier   & Training     & Training      & Classifier   \\
    \midrule
    CW-$L_2$($\kappa=40$)              &  76.4\%  &  69.1\%  &  7.9\%   &  95.7\%  &  96.5\%  &   0.8\%   \\
     PGD($\epsilon=8$, $\epsilon=0.3$) &   7.1\%  &  76.2\%  &  10.9\%  &  57.9\%  &  91.7\%  &  90.7\%   \\
    FGSM($\epsilon=0.3$)               &  72.0\%  &  71.8\%  &  17.6\%  &  98.2\%  &  98.4\%  &  90.7\%   \\
    \bottomrule
  \end{tabular}
  \end{sc}
\end{small}
\end{center}
\vskip -0.1in
\end{table*}

\subsection{Target Training performance on original, non-adversarial samples}
In Table~\ref{tbl-orig}, we show that Target Training exceeds default classifier accuracy in CIFAR10 on original, non-adversarial images when trained without adversarial samples against attacks that minimize perturbation: 86.7\% (up from 84.3\%). Furthermore, Table~\ref{tbl-orig} shows that when using adversarial samples against attacks that do not minimize perturbation, Target Training equals Adversarial Training performance.

\begin{table*}[ht]
\caption{Target Training exceeds default classifier accuracy on original, non-adversarial samples, when trained without adversarial samples against attacks that minimize perturbation in CIFAR10. Adversarial Training is not applicable because it needs adversarial samples. Target Training equals Adversarial Training performance when using adversarial samples against attacks that do not minimize perturbation.}
\label{tbl-orig}
\vskip 0.15in
\begin{center}
\begin{small}
\begin{sc}
  \begin{tabular}{lrrrrrr}
    \toprule
                 & \multicolumn{3}{c}{CIFAR10 (84.3\%)} & \multicolumn{3}{c}{MNIST (99.1\%)} \\
                                                     \cmidrule(r){2-4}   \cmidrule(r){5-7}
     Untargeted attack        & Target       & Adversarial    & Default      & Target       & Adversarial   & Default   \\                                                 
     used in training         & Training     & Training       & Classifier   & Training     & Training      & Classifier   \\
    \midrule
    none (against no                  &  86.7\%  &  NA      &  84.3\%  &  98.6\%  &  NA      &  99.1\%  \\
    perturbation attacks )            &          &          &          &          &          &          \\
    \midrule
    CW-$L_2$($\kappa=40$)             &  77.7\%  &  77.4\%  &  84.3\%  &  98.0\%  &  98.0\%  &  99.1\%  \\
    PGD($\epsilon=8$, $\epsilon=0.3$) &  76.3\%  &  76.9\%  &  84.3\%  &  98.3\%  &  98.4\%  &  99.1\%  \\
    FGSM($\epsilon=0.3$)              &  77.6\%  &  76.6\%  &  84.3\%  &  98.6\%  &  98.6\%  &  99.1\%  \\
    \bottomrule
  \end{tabular}
\end{sc}
\end{small}
\end{center}
\vskip -0.1in
\end{table*}

\subsection{Transferability Analysis}
\label{sec-transf-analysis}

For a defense to be strong, it needs to be shown to break the transferability of attacks. A good source of adversarial samples for transferability is the unsecured classifier~\cite{carlini2019evaluating}. We experiment on the transferability of attacks from the unsecured classifier to a classifier secured with Target Training. In Table~\ref{tbl-transf}, we show that Target Training breaks the transferability of adversarial samples generated by attacks that minimize perturbation much better than Adversarial Training in CIFAR10. Against the rest of attacks Target Training and Adversarial Training perform similarly.

\begin{table*}[ht]
\caption{Target Training breaks the transferability of attacks that minimize perturbation much better than Adversarial Training in CIFAR10. Against attacks that do not minimize perturbation, Target Training and Adversarial Training have comparable performance - both Target Training and Adversarial Training break the transferability of attacks in MNIST but not in CIFAR10.}
\label{tbl-transf}
\vskip 0.15in
\begin{center}
\begin{small}
\begin{sc}
  \begin{tabular}{lrrrrrr}
    \toprule
                            & \multicolumn{3}{c}{CIFAR10 (84.3\%)} & \multicolumn{3}{c}{MNIST (99.1\%)} \\
                                              \cmidrule(r){2-4}   \cmidrule(r){5-7}
    Untargeted        & Target       & Adversarial    & Default      & Target       & Adversarial   & Default   \\                                                 
    Attack            & Training     & Training       & Classifier   & Training     & Training      & Classifier   \\
    \midrule
    CW-$L_2$($\kappa=0$)              &  84.7\%  &  50.8\%  &   8.5\%  &  97.0\%  &  92.8\%  &   0.8\%  \\
    CW-$L_\infty$($\kappa=0$)         &  84.2\%  &  55.9\%  &  23.6\%  &  96.2\%  &  97.8\%  &  94.2\%  \\
    DeepFool                          &  86.6\%  &  32.3\%  &   8.6\%  &  94.9\%  &  95.9\%  &   2.8\%  \\
    \midrule
    CW-$L_2$($\kappa=40$)             &  35.8\%  &  33.8\%  &   7.9\%  &  97.8\%  &  97.9\%  &   0.8\%  \\
    PGD($\epsilon=8$, $\epsilon=0.3$) &  10.8\%  &  10.0\%  &  10.9\%  &  97.9\%  &  98.3\%  &  90.7\%  \\
    FGSM($\epsilon=0.3$)              &  34.1\%  &  45.5\%  &  17.6\%  &  72.1\%  &  75.2\%  &  90.7\%  \\
    \bottomrule
  \end{tabular}
\end{sc}
\end{small}
\end{center}
\vskip -0.1in
\end{table*}

\section{Adaptive evaluation}
\label{adapt_eval}


Many recent defenses have failed to anticipate attacks that have defeated them~\cite{carlini2019evaluating,carlini2017adversarial,athalye2018obfuscated}. Therefore, we perform an adaptive evaluation~\cite{carlini2019evaluating,tramer2020adaptive} of our Target Training defense.

\textbf{Whether Target Training could be defeated by methods used to break other defenses.} Target Training is a type of Adversarial Training because both use additional training samples, but there is no adaptive attack against Adversarial Training. Target Training uses none of previous unsuccessful defenses~\cite{carlini2016defensive,carlini2017magnet,carlini2017adversarial,athalye2018obfuscated,tramer2020adaptive} that involve adversarial sample detection, preprocessing, obfuscation, ensemble, customized loss, subcomponent, non-differentiable component. Therefore their adaptive attacks cannot be used on Target Training. In addition, we keep the loss function simple - standard softmax cross-entropy and no additional loss. Following, we discuss an adaptive attack based on the Target Training summation layer after the softmax layer.

\textbf{Adaptive attack against Target Training.} Based on the Target Training defense, we consider an adaptive attack that uses a copy of the Target Training classifier up to the softmax layer, without the last layer, to generate adversarial samples that are then tested on a full Target Training classifier. Table~\ref{tbl-adapt-attack} shows that Target Training withstands the adaptive attack.


\begin{table}[h]
\caption{Target Training withstands the adaptive attack for both CIFAR10 and MNIST. Adversarial samples are generated using a Target Training classifier up to the softmax layer, without the last layer. The generated samples are tested against the original, full Target Training classifier.}
\label{tbl-adapt-attack}
\vskip 0.15in
\begin{center}
\begin{small}
\begin{sc}
  \begin{tabular}{lrrrr}
    \toprule
     & \multicolumn{1}{c}{CIFAR10} & \multicolumn{1}{c}{MNIST} \\
     & \multicolumn{1}{c}{(84.3\%)} & \multicolumn{1}{c}{(99.1\%)} \\
                                              \cmidrule(r){2-2}   \cmidrule(r){3-3}
    Untargeted                &  Target      &  Adversarial      \\
    Adaptive Attack           &  Training      &  Training      \\
        
    
    \midrule
    CW-$L_2$($\kappa=0$)              &  84.7\%  &  97.0\%  \\ 
    CW-$L_\infty$($\kappa=0$)         &  84.2\%  &  96.3\%  \\ 
    DeepFool                          &  86.6\%  &  94.9\%  \\ 
    \midrule
    CW-$L_2$($\kappa=40$)             &  76.4\%  &  95.7\%  \\ 
    PGD($\epsilon=8$, $\epsilon=0.3$) &  76.3\%  &  92.3\%  \\ 
    FGSM($\epsilon=0.3$)              &  72.1\%  &  98.2\%  \\ 
    \bottomrule
  \end{tabular}
\end{sc}
\end{small}
\end{center}
\vskip -0.1in
\end{table}

\textbf{Iterative attacks.} The multi-step PGD~\cite{kurakin2016adversarial} attack decreases Target Training accuracy more than single-step attacks, which suggests that our defense is working correctly, according to~\citet{carlini2019evaluating}.

\textbf{Transferability.} Our transferability analysis results in Table~\ref{tbl-transf} in Subsection\ref{sec-transf-analysis} show that Target Training breaks the transferability of adversarial samples much better than Adversarial Training against attacks that minimize perturbation in CIFAR10. Target Training performance in the rest of the attacks is comparable to Adversarial Training performance. The attacks are generated with default, unsecured classifier.

\textbf{Stronger CW attack leads to better Target Training accuracy.} Increasing iterations for CW-$L_2$($\kappa=0$) 10-fold from $1K$ to $10K$ increases our defense's accuracy. In CIFAR10, the accuracy increases from $84.76$\% to $84.88$\%, in MNIST from $96.92$\% to $96.96$\%.





\section{Discussion And Conclusions}


In conclusion, we show that our white-box Target Training defense counters non-$L_\infty$ attacks that minimize perturbation in low-capacity classifiers without using adversarial samples. Target Training defends classifiers by training with duplicated original samples instead of adversarial samples. This minimizes the perturbation term in attack minimization and as a result steers attacks to non-adversarial samples. Target Training exceeds default accuracy ($84.3$\%) in CIFAR10 with $84.8$\% against CW-L$_2$($\kappa=0$), $86.6$\% against DeepFool, and $86.7$\% on original non-adversarial samples. As a form of Adversarial Training that does not use adversarial samples against attacks that minimize perturbation, Target Training defies the common justification of why Adversarial Training works. The implication is that the reason Adversarial Training works might be the same as the reason Target Training works. Not because they populate sparse areas with samples but because they steer attack convergence based on the perturbation term in attack minimization. Target Training minimizes the perturbation further than Adversarial Training and without need for adversarial samples.

\bibliography{egbib}
\bibliographystyle{icml2021}


%

\clearpage
\appendix

\balance


\vskip 0.2in
\begin{algorithm}[hbt!]
\caption{Adversarial Training of classifier $N$, based on~\cite{kurakin2016adversarial}.}
\label{alg-AT}
\begin{algorithmic}
\REQUIRE $m$ batch size, $k$ classes, $N$ classifier with $k$ output classes, ADV\_ATTACK adversarial attack, TRAIN trains classifier on a batch and labels
\ENSURE Adversarially-Trained classifier $N$
 \WHILE{training not converged}
   \STATE $B = \{x^1,..., x^m\}$\COMMENT{Get random batch}
   \STATE $G = \{y^1,..., y^m\}$\COMMENT{Get batch ground truth}
   \STATE $A = ADV\_ATTACK(N,B)$\COMMENT{Generate adv. samples from batch}
   \STATE $B' = B \bigcup A = \{x^1,..., x^m,x^1_{adv},..., x^m_{adv}\}$ \COMMENT{New batch}
   \STATE $G'=\{y^1,..., y^m,y^1,..., y^m\}$\COMMENT{Duplicate ground truth}
   \STATE TRAIN($N$, $B'$, $G'$)\COMMENT{Train classifier on new batch and new ground truth}
 \ENDWHILE
\end{algorithmic}
\end{algorithm}
\vskip -0.2in

\vskip 0.2in
\begin{algorithm}[hbt!]
\caption{Target Training of classifier $N$ using adversarial samples against attacks that do not minimize perturbation.}
\label{alg-TT-with-perturb}
\begin{algorithmic}
\REQUIRE Batch size is $m$, number of dataset classes is $k$, untrained classifier $N$ with $2k$ output classes, ADV\_ATTACK is an adversarial attack, TRAIN trains classifier on a batch and its ground truth
\ENSURE Classifier $N$ is Target-Trained against ADV\_ATTACK
 \WHILE{training not converged}
   \STATE $B = \{x^1,..., x^m\}$\COMMENT{Get random batch}
   \STATE $G = \{y^1,..., y^m\}$\COMMENT{Get batch ground truth}
   \STATE $A = ADV\_ATTACK(N,B)$\COMMENT{Generate adv. samples from batch}
   \STATE $B' = B \bigcup A = \{x^1,..., x^m,x^1_{adv},..., x^m_{adv}\}$ \COMMENT{Assemble new batch from original batch and adversarial samples}
   \STATE $G'=\{y^1,..., y^m,y^1+k,..., y^m+k\}$\COMMENT{Duplicate ground truth and increase duplicates by $k$}
   \STATE TRAIN($N$, $B'$, $G'$)\COMMENT{Train classifier on new batch and new ground truth}
 \ENDWHILE
\end{algorithmic}
\end{algorithm}
\vskip -0.2in

\begin{table}[hbt!]
\caption{Architectures of Target Training classifiers for CIFAR10 and MNIST datasets. For the convolutional layers, we use $L_2$ kernel regularizer. The pre-final Dense.Softmax layers in both models have 20 output classes, twice the number of dataset classes. The default, unsecured classifiers and the classifiers used for Adversarial Training have the same architectures, except that the softmax layer is the final layer and has only 10 outputs.} 
\label{tbl-arches}
\vskip 0.15in
\begin{center}
\begin{small}
\begin{sc}
  \begin{tabular}{ll}
    \toprule
    CIFAR10          & MNIST            \\
    \midrule
    Conv.ELU 3x3x32  & Conv.ReLU 3x3x32 \\
    BatchNorm        & BatchNorm        \\         
    Conv.ELU 3x3x32  & Conv.ReLU 3x3x64 \\
    BatchNorm        & BatchNorm        \\
    MaxPool 2x2      & MaxPool 2x2      \\
    Dropout 0.2      & Dropout 0.25     \\
    Conv.ELU 3x3x64  & Dense 128        \\
    BatchNorm        & Dropout 0.5      \\
    Conv.ELU 3x3x64  & Dense.Softmax 20 \\
    BatchNorm        & Lambda Summation 10 \\
    MaxPool 2x2      &                  \\     
    Dropout 0.3      &                  \\     
    Conv.ELU 3x3x128 &                  \\        
    BatchNorm        &                  \\     
    Conv.ELU 3x3x128 &                  \\
    BatchNorm        &                  \\
    MaxPool 2x2      &                  \\
    Dropout 0.4      &                  \\ 
    Dense.Softmax 20 &                  \\
    Lambda Summation 10 &               \\
    \bottomrule
  \end{tabular}
\end{sc}
\end{small}
\end{center}
\vskip -0.1in
\end{table}

\begin{table*}[hbt!]
\caption{Here, we show that targeted attacks are not strong against default classifiers, decreasing default accuracies very little. Target Training and Adversarial Training have roughly equal performance against targeted attacks, except for CW-$L_{\infty}$ in CIFAR10 where Target Training has better accuracy, and PGD where Adversarial Training has better accuracy. DeepFool attacks are not applicable because they cannot be targeted.}
\label{tbl-full-targ}
\vskip 0.15in
\begin{center}
\begin{small}
\begin{sc}
  \begin{tabular}{lrrrrrr}
    \toprule
                            & \multicolumn{3}{c}{CIFAR10 (84.3\%)} & \multicolumn{3}{c}{MNIST (99.1\%)} \\
                                                 \cmidrule(r){2-4}   \cmidrule(r){5-7}    
    Targeted          & Target       & Adversarial    & Default      & Target       & Adversarial   & Default   \\                                                 
    Attack            & Training     & Training       & Classifier   & Training     & Training      & Classifier   \\
    
    \midrule
    CW-$L_2$($\kappa=0$)              &  82.8\%  &  83.1\%  &  84.0\%  &  96.9\%  &  98.9\%  &  98.3\%  \\
    CW-$L_{\infty}$($\kappa=0$)       &  69.0\%  &  50.1\%  &  72.7\%  &  98.2\%  &  98.1\%  &  98.6\%  \\
    DeepFool                          &  NA      &  NA      &  NA      &  NA      &  NA      &  NA      \\
    \midrule
    CW-$L_2$($\kappa=40$)             &  84.4\%  &  84.1\%  &  85.7\%  &  99.0\%  &  99.0\%  &  99.0\%  \\
    PGD($\epsilon=8$, $\epsilon=0.3$) &  21.4\%  &  34.5\%  &  44.9\%  &  85.3\%  &  97.6\%  &  96.4\%  \\
    FGSM($\epsilon=0.3$)              &  77.9\%  &  80.5\%  &  46.4\%  &  98.2\%  &  98.4\%  &  96.4\%  \\
    \bottomrule
  \end{tabular}
\end{sc}
\end{small}
\end{center}
\vskip -0.1in
\end{table*}


\end{document}